\documentclass[sigconf]{article}
\usepackage{microtype}
\usepackage{graphicx}
\usepackage{enumitem}
\usepackage{subfigure}
\usepackage{multirow}
\usepackage{float}
\usepackage{hyperref}
\usepackage{balance}
\usepackage{flushend}
\usepackage{adjustbox}
\usepackage{colortbl}
\usepackage{algorithm} 
\usepackage{algorithmic}  
\usepackage[algo2e]{algorithm2e} 
\usepackage[utf8]{inputenc}
\usepackage[T1]{fontenc}    
\usepackage{multirow}
\usepackage{hyperref}       
\usepackage{url}            
\usepackage{booktabs}       
\usepackage{nicefrac}       
\usepackage{microtype}      
\usepackage{amsthm}
\usepackage{amsmath}
\usepackage{amssymb}
\usepackage{amsfonts}       
\usepackage{mathtools}
\usepackage{mathrsfs}
\usepackage{xcolor}
\usepackage{relsize}    
\usepackage{graphicx}
\usepackage{enumerate}
\usepackage{pgfplots}
\usepackage{filecontents}
\usepackage{multicol}
\usepackage{tikz}
\usepackage{environ}
\usepackage[symbol]{footmisc}
\usepackage{pgfplots}
\usetikzlibrary{3d}
\usepackage{wrapfig}
\usepackage{tikz-3dplot}
\usepgfplotslibrary{patchplots}
\usetikzlibrary{arrows}
\usepgfplotslibrary{fillbetween}
\usepackage{blindtext}
\usepackage{makecell}
\usepackage[font=scriptsize]{caption}

\usepackage{sidecap}

\SetCommentSty{mycommfont}
\SetKwComment{Comment}{$\blacktriangleleft$\ }{$\blacksquare$}
\SetKwComment{tcc}{$\blacktriangledown$\ }{$\blacktriangledown$}
\SetKwInput{KwInput}{Input}                
\SetKwInput{KwOutput}{Output}              
\SetKwInput{KwParameter}{Parameter}              

\pgfplotsset{compat=1.16} 
\newcommand{\executeiffilenewer}[3]{%
\ifnum\pdfstrcmp{\pdffilemoddate{#1}}%
{\pdffilemoddate{#2}}>0%
{\immediate\write18{#3}}\fi%
}

\usepackage[accepted]{icml2023}
\icmltitlerunning{Dynamic Switch Layers For Unsupervised Learning}

\begin{document}
\twocolumn[
\icmltitle{Dynamic Switch Layers For Unsupervised Learning}
\icmlsetsymbol{equal}{*}
\begin{icmlauthorlist}
\icmlauthor{Haiguang Li}{google}
\icmlauthor{Usama Pervaiz}{google}
\icmlauthor{Michał Matuszak}{google}
\icmlauthor{Robert Kamara}{google}
\icmlauthor{Gilles Roux}{google}
\icmlauthor{Trausti Thormundsson}{google}
\icmlauthor{Joseph Antognini}{google}
\end{icmlauthorlist}
\icmlaffiliation{google}{Google LLC, Mountain View, CA 94043, USA}
\icmlcorrespondingauthor{Haiguang Li}{haiguang@google.com}
\icmlkeywords{Machine Learning, ICML}
\vskip 0.3in
]

\printAffiliationsAndNotice{}

\begin{abstract}
On-device machine learning (ODML) enables intelligent applications on resource-constrained devices. However, power consumption poses a major challenge, forcing a trade-off between model accuracy and power efficiency that often limits model complexity. The previously established Gated Compression (GC) layers offer a solution, enabling power efficiency without sacrificing model performance by selectively gating samples that lack signals of interest. However, their reliance on ground truth labels limits GC layers to supervised tasks.
This work introduces the Dynamic Switch Layer (DSL), extending the benefits of GC layers to unsupervised learning scenarios, and maintaining power efficiency without the need for labeled data. The DSL builds upon the GC architecture, leveraging a dynamic pathway selection, and adapting model complexity in response to the innate structure of the data. 
We integrate the DSL into the SoundStream architecture and demonstrate that by routing up to 80\% of samples through a lightweight pass we achieve a 12.3x reduction in the amount of computation performed and a 20.9x reduction in model size. This reduces the on-device inference latency by up to 26.5\% and improves power efficiency by up to 21.4\% without impacting model performance.
\end{abstract}

\section{Introduction}
\begin{figure}[tb!]
    \centering
    \includegraphics[width=\linewidth]{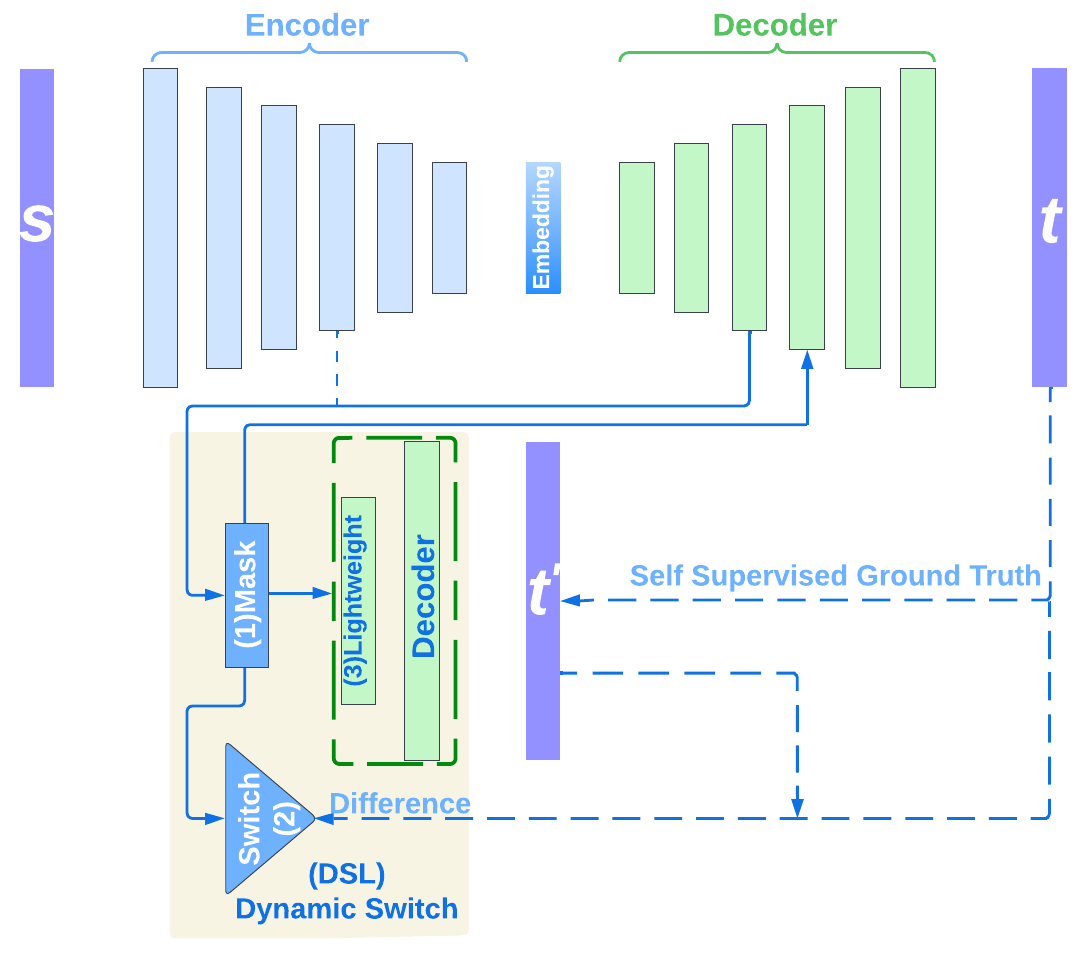}\vspace{-8pt}
    \caption{The Proposed \textit{Dynamic Switch Layer (DSL)}: \textit{(1)Mask} for activation sparsity, \textit{(2)Switch} for path switching/routing, and \textit{(3)Lightweight Decoder} for output generating from the lightweight branch.}
    \label{fig:DSL}\vspace{-12pt}
\end{figure}
On-device machine learning (ODML) has emerged as an powerful tool for improving user experiences, preserving data privacy, and reducing reliance on cloud infrastructures \cite{chen2014diannao}. Implementing machine learning (ML) models on a broad spectrum of devices, from smartphones to wearables and IoT devices, enables real-time, context-sensitive, and personalized user interactions. However, the deployment of these  models on resource-constrained devices poses significant power consumption challenges that must be addressed for the advancement of ODML applications \cite{han2015deep, EdgeComputing:AReview2018}.

Power consumption in ML models is influenced by a multitude of factors, such as model architecture, computational load, memory footprint, and frequency of memory access. These factors result in issues like shortened battery life, performance trade-offs, lagged responsiveness, thermal issues, and restricted functionality \cite{chen2016eyeriss, gupta2015deep}. Addressing these limitations by devising power-efficient methods for ODML is therefore crucial.

Researchers have proposed various techniques to mitigate these power consumption hurdles in ODML, including model compression, hardware optimization, and neural architecture search (NAS) \cite{han2015deep, hinton2015distilling, zhou2017incremental, chen2016eyeriss, tan2019mnasnet}. The integration of these techniques has led to the creation of more efficient heterogeneous hardware architectures using specialized cores to offer optimised pathways for low-power ML inference.

To further improve efficiency, we recently introduced Gated Compression (GC) layers \cite{li2023gcl}. GC layers transform existing neural networks into Gated Neural Networks, optimizing them for improved on-device performance by reducing power consumption, boosting accuracy, and enabling better utilization of heterogeneous hardware. However, the initial design of GC layers were suited for supervised learning as they depended on labeled data for their gating mechanism.

This work introduces the novel Dynamic Switch Layer (DSL) by generalizing the GC layer into a dynamic architecture. The DSL offers the following key properties:
\begin{itemize}
    \item \textbf{Unsupervised Learning:} The DSL operates without the need for ground truth labels by learning sparsification and path switching/routing strategies directly from the data. This extends its benefits to unsupervised learning scenarios.
    \item \textbf{Activation Sparsity:} The DSL learns to induce sparsity within intermediate feature maps. This minimizes data transfer, reducing computational overhead, which is critical in heterogeneous computing systems where data transmission is costly.
    \item \textbf{Distributed Models:} The DSL acts as natural delimiters for large-scale models, facilitating distribution across varied computational resources (across cores or extending to the cloud). This, combined with sparsity and path switching, enables the use of more powerful models while maintaining low power consumption.
    \item \textbf{Generality:} The DSL demonstrates superior generalization by subsuming the supervised GC layer as a special case. This highlights its ability to excel in both supervised and unsupervised learning settings.
\end{itemize}

Our proposed DSL draws inspiration from the human brain, much like how sparsity in brain networks ensures that only essential connections are maintained to conserve energy \cite{attwell2001energy, pervaiz2020optimising}. This is paralleled to how activation sparsity is applied in the DSL to reduce computational load. Additionally, our brain is adept at allocating its computational power in a highly efficient manner, assigning more mental resources to complex tasks while conserving energy during simpler activities \cite{causse2017mental}. The dynamic allocation of resources in human brain is similar to our proposed switching mechanism in the DSL, channeling greater processing power for intricate computations while conserving energy during less demanding tasks.

The contributions of this work are as follows:
\begin{itemize}
    \item We propose the DSL, equipped with an effective loss function. This enables fine-tuning of on-device models for network path switching/routing, activation sparsity, and overall performance, leading to substantial efficiency gains.
    \item We show through a variety of experiments on audio datasets that our DSL can be seamlessly integrated into the SoundStream model \cite{zeghidour2021soundstream}.  This improves power efficiency by routing up to 80\% of samples to a lightweight pass, achieving a 20.9x size reduction and 12.3x computation reduction while maintaining both model accuracy and downstream application performance.
    \item We show that models supported by our DSL demonstrate significantly reduced on-device inference latency (up to 26.5\%) and improved power efficiency (up to 21.41\%) without compromising the models' performance, showcasing its practicality in real-world applications.
\end{itemize}

\section{Dynamic Switch Layers For Unsupervised Learning}

As elaborated in \cite{li2023gcl}, the original GC layer serves as a specific gating mechanism, filtering out samples devoid of the signal of interest (negative samples), and compressing irrelevant or superfluous feature dimensions in samples containing positive signals.

\cite{li2023gcl} showed that the original GC layer exhibits robust performance for always-on supervised classification tasks. Beyond enhancing the baseline model's efficiency, it substantially trims power consumption. To perform efficient early stopping of negative samples, the original class labels are converted into positive and negative classifications via the $\Omega(\cdot)$ remapping function. When a sample is definitively deemed negative by the gate, the neural network immediately exits and bypasses the remaining layers. Despite these merits, the original GC layer is unsuitable for unsupervised tasks, as the $\Omega(\cdot)$ function requires ground truth class data to learn to discriminate negative from positive samples. To address this limitation, we introduce the Dynamic Switch Layer (DSL), designed specifically for unsupervised learning.

\subsection{Unsupervised Learning with Autoencoders}

Unsupervised learning is a branch of ML where algorithms discover patterns within unlabeled data. Unlike supervised learning, which relies on labeled examples, unsupervised methods must uncover structure and relationships solely by analyzing the data itself. This domain has diverse applications including representation learning, noise reduction, and generative modeling.

Autoencoders are a key unsupervised learning technique, designed to learn efficient data representations through a dual process of compression and reconstruction of input data. An autoencoder consist of two primary components:
\begin{itemize}
    \item\textit{Encoder:} This segment of the neural network compresses input data into a condensed, lower-dimensional form known as the latent space. This process can be seen as feature extraction or encoding.
    \item\textit{Decoder:} Following the encoder, this neural network attempts to reconstruct the original input data as accurately as possible from the compressed latent representation. This process is known as decoding or reconstruction.
\end{itemize}

During training, the autoencoder's goal is to minimize the difference between the original input and its reconstruction. Standard loss functions include mean squared error or cross-entropy. Through training, the autoencoder learns to identify and compress the most important features of the data. This creates a compact, informative representation that has numerous downstream applications.

\subsection{Dynamic Switch Layer}

Figure \ref{fig:DSL} shows how to integrate the DSL into an existing unsupervised autoencoder. The DSL can integrate seamlessly with the baseline model without requiring further adjustments.

\subsubsection{Main Components}
The DSL, as depicted in Figure \ref{fig:DSL}, has three primary components:
\begin{itemize}
\item \textbf{(1)Mask} ($\mathcal{M}$): compresses the activation feature map, reducing data volume for efficiency while preserving essential information.
\item \textbf{(2)Switch} ($\mathcal{S}$): serves as the decision-making hub, dynamically directing simpler samples to a less resource-intensive pathway (i.e., the lightweight decoder in Figure \ref{fig:DSL}), thus saving computational resources.
\item \textbf{(3)Lightweight Decoder} ($\mathcal{D}$): A unique component of the DSL that differentiates itself from the gating mechanisms in the original GC layer \cite{li2023gcl} by redirecting rather than terminating the signal flow, allowing for computation along a low-power pathway. In contrast to the gates \cite{li2023gcl} that rely on ground truth labels to determine signal flow interruption (i.e., negative class), the lightweight decoder uses data features to make switching decisions. This is essential for unsupervised learning as it eliminates the need for labeled data.
\end{itemize}

While the Switch, analogous to the gating mechanisms in GC layers, maintains its original role of routing signals (to lightweight/low-power pathway as depicted in Figure \ref{fig:DSL}), and mask retains functionality of compressing the activation maps, the lightweight decoder emerges as the main improvement of the DSL over the GC layer. It equips the DSL to handle unsupervised tasks with reduced computational requirements by processing less complex operations through a low-power route.

\subsubsection{The DSL Loss Function}

We now turn to exclusively on the loss of the newly inserted DSL. Let $x$ represent the input sample, $\mathcal{F}^{\mapsto i}(\cdot)$ denote the operation performed by layers up to the layer immediately preceding the DSL, and $\mathcal{F}^{i+1\mapsto}(\cdot)$ signify the operation executed by layers following the DSL. Then, we have
\begin{equation}
    \left\{\begin{aligned}
f_m(x)  &= \mathcal{M}\left(\mathcal{F}^{\mapsto i}(x)\right)\\ 
\mathcal{S}(f_m(x)) &= \mathcal{S}\left(\mathcal{M}\left(\mathcal{F}^{\mapsto i}(x)\right)\right) \\
\mathcal{D} \left(f_m(x)\right) &= \mathcal{D} \left(\mathcal{M}\left(\mathcal{F}^{\mapsto i}\left(x\right)\right)\right) \\
\mathcal{F}^{i+1\mapsto} \left(f_m(x)\right)  &=  \mathcal{F}^{i+1\mapsto }\left(\mathcal{M}\left(\mathcal{F}^{\mapsto i}\left(x\right)\right)\right)
\end{aligned}\right.
\end{equation}
where $f_m(x)$, $\mathcal{S}(f_m(x))$, and $\mathcal{D} \left(f_m(x)\right)$ represent the outputs of the mask ($\mathcal{M}$), Switch ($\mathcal{S}$), and lightweight decoder ($\mathcal{D}$) of the DSL, respectively. Meanwhile, $\mathcal{F}^{i+1\mapsto} \left(f_m(x)\right)$ corresponds to the output of the end-to-end autoencoder, which serves as the ground truth for optimizing the DSL.

\textbf{Compression Loss.} The mask ($\mathcal{M}$) is identical to the compression layer in the GC layer, and its loss can be computed as:
\begin{equation}\label{eq:compression_loss}
    \mathcal{L}_\text{comp}(\mathcal{W_M}) = \left \| \mathcal{W_M} \right \|_\textrm{c},
\end{equation}
where $\left \| \cdot \right \|_\textrm{c}$ is a sparsity regularization term (e.g. $\mathcal{L}_1$) applied to the weight matrix $\mathcal{W_M}$ to promote sparsity in the activation outputs.

\textbf{Switch Loss.} Similar to the gate in the original GC layer, the Switch $\mathcal{M}$ is designed to dynamically switch to the lightweight path for the easier samples. Its loss can be computed as:
\begin{equation}\label{eq:gate_loss}
\begin{aligned}
    \mathcal{L}_\text{Switch}(x)&=\left\|
    \mathcal{D}\left(f_m(x)\right) - \mathcal{F}^{i+1\mapsto }\left(f_m(x)\right)\right\|_\textrm{Switch} ,
\end{aligned}
\end{equation}
where $f_{m}(x)$ represents the output after applying the mask ($\mathcal{M}$) to the transformed input $x$ using $\mathcal{F}^{\mapsto i}$. The term  $\left\|\cdot\right\|_\textrm{Switch}$ denotes a loss function that measures the difference between the outputs of the full and lightweight decoders.

\textbf{Lightweight Decoder Loss.} The lightweight decoder ($\mathcal{D}$) is optimized to mimic the original decoder's behavior by minimizing their output differences.
Its loss can be computed as:
\begin{equation}\label{eq:lightweight_decoder_loss}
    \mathcal{L}_\text{lwd}(x)=\mathcal{L}\left (\mathcal{D}\left(f_m(x)\right), \mathcal{F}^{i+1\mapsto }\left(f_m(x)\right)\right ) ,
\end{equation}
where $\mathcal{L}\left( \cdot, \cdot \right )$ is a loss function (e.g. cross entropy loss).

\textbf{Overall Loss of DSL}. The overall loss of an DSL can be computed by the weighted sum of the compression loss (Equation \ref{eq:compression_loss}), Switch loss (Equation \ref{eq:gate_loss}), and lightweight decoder loss (Equation \ref{eq:lightweight_decoder_loss}):
\begin{equation}\label{eq:ssl_gcl_overall}
    \mathcal{L}_\text{dsl}(x) = \alpha  \mathcal{L}_\text{Switch}(x) + \alpha \mathcal{L}_\text{lwd}(x) + \beta \mathcal{L}_\text{comp}(\mathcal{W_M}),
\end{equation}
where $\alpha$ and $\beta$ are hyper-parameters that allow for fine-grained tuning of the corresponding terms efficiently.
From Equations \ref{eq:gate_loss} and \ref{eq:lightweight_decoder_loss}, both $\mathcal{L}_\text{gate}(x)$ and $\mathcal{L}_\text{lwd}(x)$ share the same origin, consequently, they are regulated by the same hyper-parameter $\alpha$.

With Equation \ref{eq:ssl_gcl_overall}, the loss of a newly inserted DSL can be combined and trained together with the new model in an end-to-end fashion.

\subsubsection{Design of the Lightweight Decoder}
The lightweight decoder must maintain compatibility with the input and output shapes of the original decoder while using a smaller number of parameters in the intermediate layers. A typical strategy is to mirror the original decoder's architecture with reduced complexity. For  convolutional neural networks (CNNs), this means fewer kernels per layer; for attention-based models, fewer attention heads. This provides a baseline for performance, and the lightweight decoder can then be further optimized and fine-tuned as needed.

\subsubsection{A Generalization of the GC Layer}
The DSL is a generalization of the GC layer. The core distinctions are: (1) the DSL includes a lightweight decoder tailored for unsupervised scenarios, and (2) it forgoes reliance on ground truth labels, making it applicable to both supervised and unsupervised learning. In always-on ambient settings where the focus is on detecting positive signal events, a lightweight decoder is unnecessary when a sample is negative. Additionally, supervised learning naturally provides ground truth labels, making them a preferable supervisory signal. Therefore, the GC layer can be considered a special case of the DSL, specifically designed for always-on ambient computing with supervised learning.

\subsubsection{Related Work}
\balance
\textbf{Knowledge Distillation} \cite{Distilling2015} is a technique to transfer knowledge from a large, complex model (the teacher) to a smaller, more efficient one (the student). Within the DSL (Figure \ref{fig:DSL}), the lightweight decoder functions as the student, trained to mimic the behavior of the original decoder (the teacher). This allows the lightweight decoder to achieve comparable performance with a significantly reduced size. Additionally, the Switch component of the DSL learns to assess the performance of lightweight decoder. If the output difference between the lightweight decoder and original decoder is minimal (Equation \ref{eq:gate_loss}), this indicates high confidence on the lightweight decoder's output. During inference, the Switch can selectively route samples to the lightweight decoder when confidence is high, optimizing computation for improved efficiency without compromising performance.

\textbf{Contrastive Learning} without negative samples \cite{grill2020BYOL} is a self-supervised approach that learns robust data representations by focusing on similarities rather than explicit contrasts. The goal is to make models insensitive to minor variations, emphasizing core characteristics of the data. Bootstrap Your Own Latent (BYOL) \cite{grill2020BYOL} is a powerful technique that distinguishes itself by eliminating the need for comparing negative pairs, unlike traditional contrastive learning approaches. It trains a model by predicting one augmented view of an input from another using a Siamese network. The objective is to minimize the difference between these views. The DSL (Figure \ref{fig:DSL}) adopts a similar philosophy but extends it further by not only minimizing the difference between the outputs of the original and lightweight decoders but by further assessing the confidence in lightweight decoder's performance.

\textbf{Masked Auto Encoders (MAEs)} are a variant of unsupervised autoencoders \cite{he2022mae}. In a MAE, a portion of the input data is masked or removed during training, forcing the model to reconstruct the original input from this incomplete information. This encourages the model to learn robust representations that capture the underlying structure of the data. The DSL draws inspiration from this masking concept. While MAE masks the input directly, the DSL applies masking to the intermediate latent representations (or embeddings).

\section{Experiments}

In this section, we apply the DSL for tokenization tasks within the speech domain. We begin by detailing the datasets used, the evaluation protocols, and implementation specifics. This is followed by a discussion of the results. Subsequently, we conduct an ablation study to measure the impact of the various components of the DSL and share insights derived from our findings.

\subsection{Datasets, Architecture, and Evaluation Details}

\subsubsection{Datasets}
This work uses two publicly available datasets: LibriTTS \cite{zen2019libritts} for model training and AudioSet \cite{jort_audioset_2017} for testing. LibriTTS is a large-scale, high-quality dataset specifically designed for text-to-speech (TTS), and offers cleaner audio, diverse speaking styles, and more accurate text-audio alignment compared to its predecessor, Librispeech \cite{Librispeech2015}. AudioSet's vast collection of labeled audio events (632 classes), spanning human, animal, musical, and environmental sounds, provides a robust testbed for evaluating our model's ability to handle real-world audio complexity.

\textit{Downstream Use Case Evaluation:} To assess performance in downstream tasks, we utilize two specialized datasets:  Speech Commands \cite{warden2018speech} and LibriCount \cite{stoter2018libricount}. Speech Commands, with its variety of spoken words and background noise, is ideal for evaluating our model's keyword spotting and small-vocabulary recognition capabilities. LibriCount, a  synthetic dataset simulating cocktail party environments with multiple speakers, allows us to specifically study the model's performance in speaker count estimation.

\subsubsection{Model Architecture}

SoundStream \cite{zeghidour2021soundstream} introduces an end-to-end neural audio codec that directly generates a bitstream from raw audio using deep learning, outperforming traditional codecs in quality while maintaining comparable compression.

\begin{figure}[tbh!]
  \centering 
  \includegraphics[width=1\linewidth]{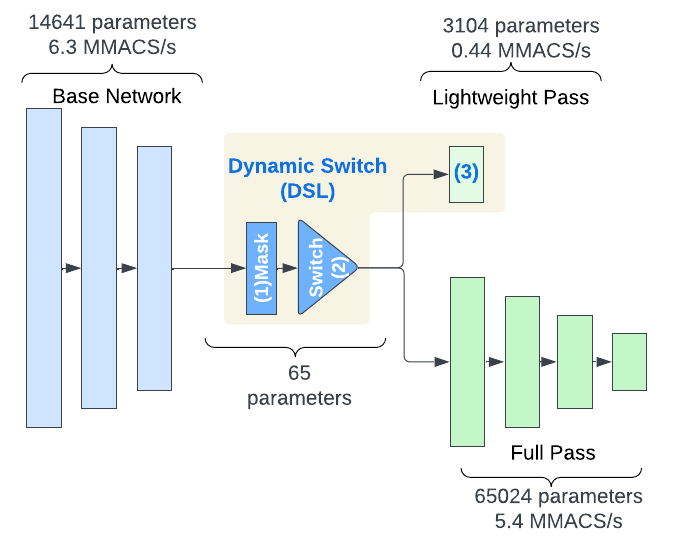}
  \vspace{-12pt}
\caption{Power-Efficiency Enhancement in SoundStream Encoder with DSL Integration. The diagram illustrates the reduction in parameter count and computational load when employing the lightweight pass.}
  \label{fig:dsl_encoder}
\end{figure}

The SoundStream codec compromises a neural encoder, a neural decoder, and a quantization module in between. In real-world applications, continuous encoder operation on resource-limited devices is essential. Our experiments specifically adapt the SoundStream encoder, applying the DSL to significantly improve its power efficiency (Figure \ref{fig:dsl_encoder}). Importantly, we maintain end-to-end training of the updated encoder alongside the decoder and quantization module to ensure optimal performance. Moreover, Figure \ref{fig:dsl_encoder} highlights another crucial aspect: the role of the Switch within the DSL. The ability of the Switch to accurately predict when the lightweight pass will closely match the original encoder's output is essential for maintaining performance while improving efficiency. 

The original version of SoundStream encoder, as presented in \citet{zeghidour2021soundstream}, had a parameter count of more than half a million. We restructured it into a leaner model with approximately 70k parameters by reducing the number of residual blocks and adjusting the kernel size. This optimization maintained SoundStream's performance with no more than a 3\% drop in audio quality. Subsequently, within this optimised SoundStream encoder, we introduced a lightweight path. This path was achieved by scaling down the number of filters in specific layers, thereby reducing the computational load while preserving the original input/output structure of the model. 

As illustrated in Figure \ref{fig:dsl_encoder}, the DSL enables a dramatic reduction in sub-model size by routing easy samples to the lightweight pass. This reduces size from 65,024 to 3104, a 20.9x improvement.  Additionally, the computational load is slashed by 12.3x, from 5.4 Million Multiply-Accumulates per Second (MMACS/s) to 0.44 MMACS/s.

\subsubsection{Evaluation}
Virtual Speech Quality Objective Listener (ViSQOL) \cite{ViSQOL, ViSQOLAudio, chinen2020visqol} is an objective, full-reference metric for evaluating perceived audio quality. It employs a spectro-temporal similarity measure between a reference and a test speech signal to generate a MOS-LQO (Mean Opinion Score - Listening Quality Objective) score. MOS-LQO scores range from 1 (worst) to 5 (best).

We implemented all methods using TensorFlow 2.x \citep{tensorflow2015-whitepaper}, and used the Adam optimizer \citep{adom2015} for model training.

\subsection{DSL Performance}
This section details a series of experiments designed to thoroughly assess the DSL's performance.

\begin{figure*}
    \centering
    \includegraphics[width=0.98\textwidth]{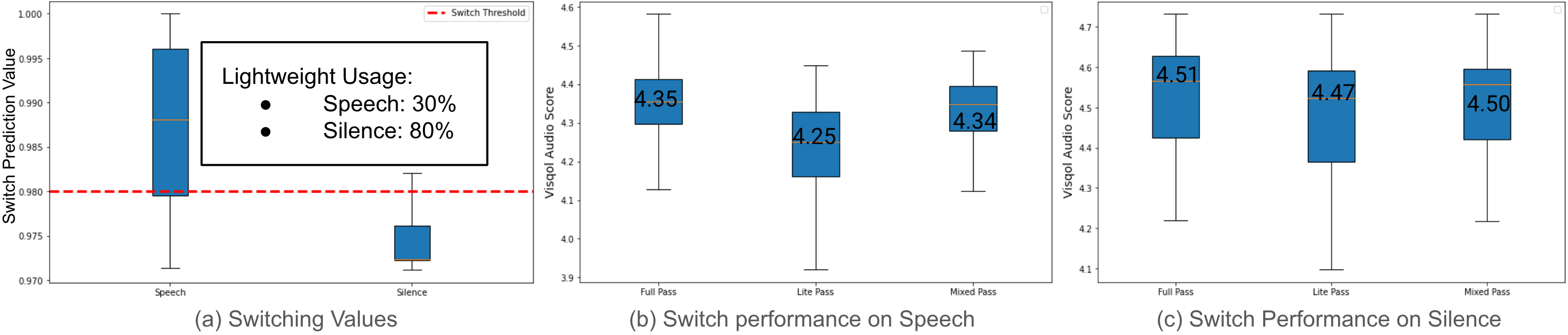}
    \caption{Switch Performance for Dynamic Routing: (a) Switch predicted values on speech data and silence data, (b) Switch performance by routing speech samples to different passes, and (c) Switch performance by routing silence samples to different passes.}
    \label{fig:dsl_switch_overall}
\end{figure*}
\begin{figure*}
    \centering
    \includegraphics[width=0.98\textwidth]{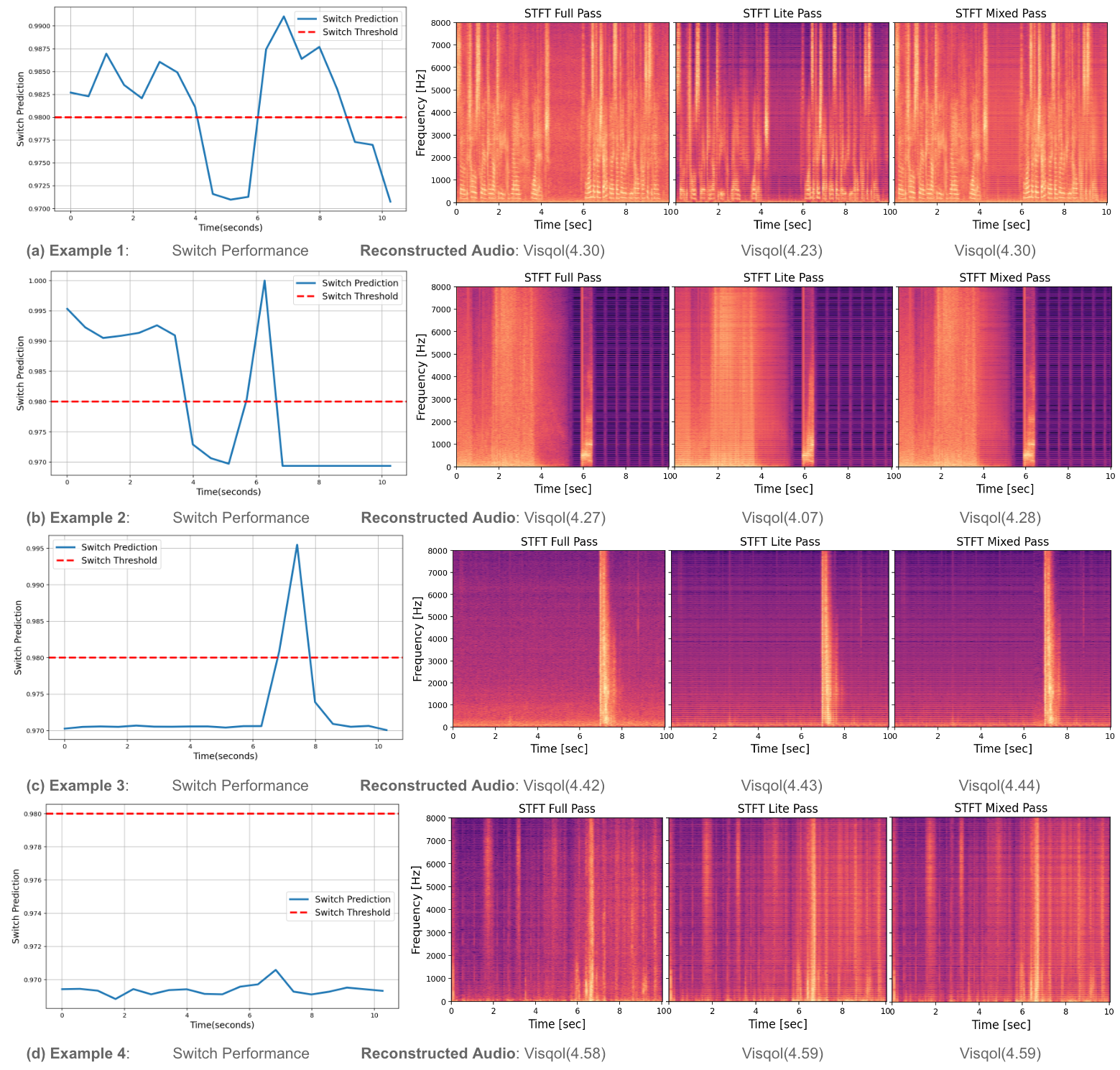}
    \caption{Dynamic Routing of Audio Segments in DSL. Each example shows the Switch's predictive behaviour over time (left) and the corresponding spectrograms with Visqol scores for full, lightweight and mixed pass output (right). }
    \label{fig:qualitative}
\end{figure*}

\begin{table*}[]
\caption{On Device Inference Time and Power Consumption Performance}
\centering
\label{tab:ondevice}
\begin{adjustbox}{max width=0.97\textwidth}

\begin{tabular}{|l|c|cccc|cccc|}
\hline
\rowcolor[HTML]{EFEFEF} 
\multicolumn{1}{|c|}{\cellcolor[HTML]{EFEFEF}}                                                                            & \cellcolor[HTML]{EFEFEF}                                                                                      & \multicolumn{4}{c|}{\cellcolor[HTML]{EFEFEF}Inference Time (ms)}                                                                                                             & \multicolumn{4}{c|}{\cellcolor[HTML]{EFEFEF}Power Consumption (mW)}                                                                                                          \\ \cline{3-10} 
\rowcolor[HTML]{EFEFEF} 
\multicolumn{1}{|c|}{\multirow{-2}{*}{\cellcolor[HTML]{EFEFEF}\begin{tabular}[c]{@{}c@{}}\end{tabular}}} & \multirow{-2}{*}{\cellcolor[HTML]{EFEFEF}\begin{tabular}[c]{@{}c@{}}Lightweight \\ Pass Usage\%\end{tabular}} & \multicolumn{1}{c|}{\cellcolor[HTML]{EFEFEF}Lightweight} & \multicolumn{1}{c|}{\cellcolor[HTML]{EFEFEF}Full} & \multicolumn{1}{c|}{\cellcolor[HTML]{EFEFEF}Mixed} & Saving\% & \multicolumn{1}{c|}{\cellcolor[HTML]{EFEFEF}Lightweight} & \multicolumn{1}{c|}{\cellcolor[HTML]{EFEFEF}Full} & \multicolumn{1}{c|}{\cellcolor[HTML]{EFEFEF}Mixed} & Saving\% \\ \hline
\cellcolor[HTML]{EFEFEF}Silence                                                                                           & 80\%                                                                                                          & \multicolumn{1}{c|}{}                                    & \multicolumn{1}{c|}{}                             & \multicolumn{1}{c|}{10.84}                         & 26.5\%   & \multicolumn{1}{c|}{}                                    & \multicolumn{1}{c|}{}                             & \multicolumn{1}{c|}{9.09}                          & 21.4\%   \\ \cline{1-2} \cline{5-6} \cline{9-10} 
\cellcolor[HTML]{EFEFEF}Speech                                                                                            & 30\%                                                                                                          & \multicolumn{1}{c|}{\multirow{-2}{*}{9.86}}              & \multicolumn{1}{c|}{\multirow{-2}{*}{14.74}}      & \multicolumn{1}{c|}{13.28}                         & 9.9\%    & \multicolumn{1}{c|}{\multirow{-2}{*}{8.47}}              & \multicolumn{1}{c|}{\multirow{-2}{*}{11.56}}      & \multicolumn{1}{c|}{10.63}                         & 8.0\%    \\ \hline
\end{tabular}
\end{adjustbox}
\end{table*}

\begin{figure*}
  \centering
  \includegraphics[width=.98\linewidth]{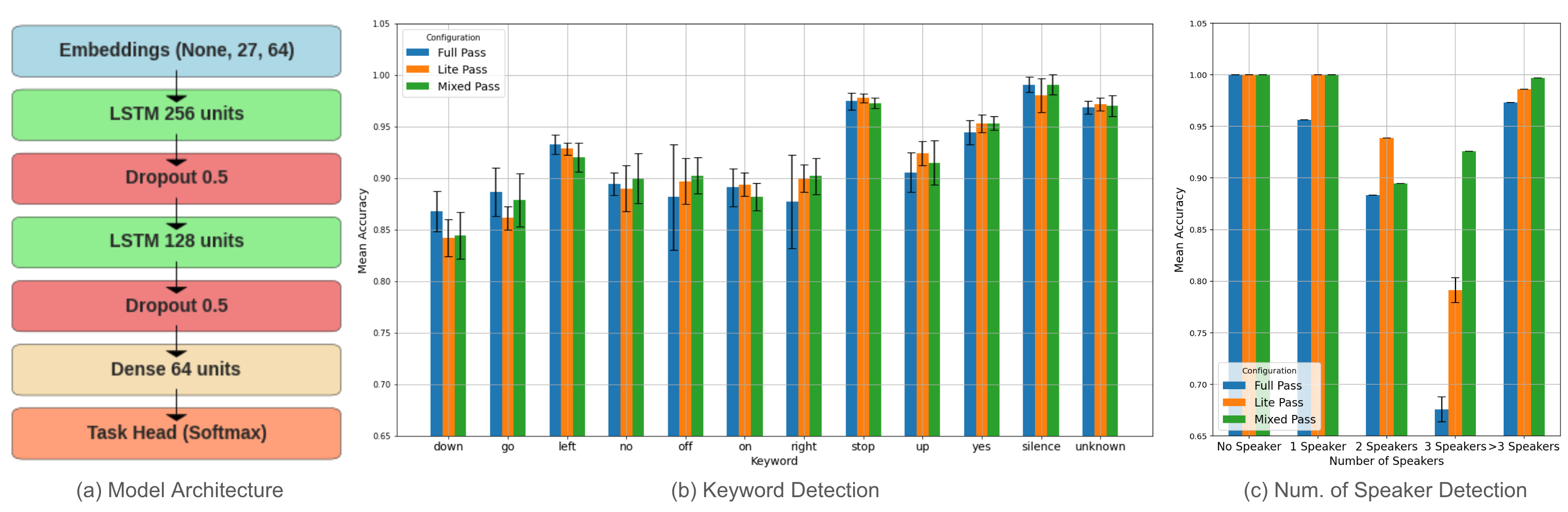}
\caption{Performance for Downstream Use-Cases in Keyword and Speaker Detection. Performance comparison across original full pass, lightweight pass, and the DSL's mixed pass, highlighting the efficiency of dynamic routing in maintaining or enhancing audio quality.}
  \label{fig:downstrea-usecases}
\end{figure*}

\subsubsection{Overall Performance}

In real-world applications, the DSL Switch dynamically determines the routing of audio segments, resulting in a mixed pass that combines the lightweight and full pass outputs. When the Switch predicts lower differences (i.e., below the threshold), indicating simpler audio content, it routes the sample through the lightweight pass. This selective routing is critical because it enables resource conservation without significantly degrading performance. In Figure \ref{fig:dsl_switch_overall}, we analyze the Switch performance on the AudioSet data, for both ``silence'' and ``speech'' class. In the AudioSet, data under the ``silence'' label included not only pure silence but also quiet tones and ambient noises.

To optimize the switching mechanism, we analyzed the Switch's predicted values for ``silence'' (easy) and ``speech'' (hard) audio samples offline. Figure \ref{fig:dsl_switch_overall}(a) confirms that, as expected, ``silence'' samples have lower predicted differences, making them suitable for the lightweight pass. Additionally, by setting the Switch threshold at 0.98, we achieve a significant efficiency gain:
(1) approximately 80\% of ``silence'' samples are routed to the lightweight pass;
(2) around 30\% of ``speech'' samples also use the lightweight pass.

The performance metrics (Visqol scores) in Figure \ref{fig:dsl_switch_overall} indicate that the quality of mixed pass remains robust against the full pass, showing that the DSL approach yields high efficiency with negligible compromises on the audio output quality. Figure \ref{fig:dsl_switch_overall}(b) and (c) demonstrate this success: Visqol scores remain consistent for both ``silence'' (4.50 mixed pass vs. 4.51 full pass) and ``speech'' samples (4.34 mixed pass vs. 4.35 full pass), confirming that the mixed pass strategy maintains audio quality.

\subsubsection{Qualitative Examples}
Figure \ref{fig:qualitative} demonstrates the DSL's ability to dynamically route audio segments to either the lightweight or full pass, based on the Switch's predictions.

This adept routing allows the reconstructed audio to achieve the same or even better quality compared to using the full path exclusively. Examples include: 
(1) In Example 1, using the mixed pass strategy maintains audio quality (Visqol 4.30), similar to exclusively using the full path (Visqol 4.30), revealing that lightweight output can match the full path's performance for less complex segments.
(2) In Examples 2, 3, and 4, performance of mixed pass routing surpasses the full path's performance. 

There is also another critical observation from these examples (i.e., Example 1 and Example 2) that relying solely on the lightweight pass leads to a notable decline in performance.

\subsubsection{On Device Inference Latency and Power Saving}
Here we assess the real-world impact of our DSL on the entire encoder model (including base network, full/lightweight passes as depicted in Figure \ref{fig:dsl_encoder}).  For this evaluation, we deployed the model on a device equipped with an A32 core. Our focus was on two key metrics: inference time and power consumption.

To accurately measure power consumption, a hardware development board is used for providing external access to power rails. We calculated the model's power usage by taking measurements with and without it performing  inference and subtract the two. Each measurement lasted 30 seconds, and we repeated this process ten times for consistency. Additionally, we employed a software profiler to track the time taken for each inference.

Using audio segments from the AudioSet dataset for benchmarking, we observed significant reductions in inference latency (9.9\% for speech clips and 26.5\% for silent clips) and improved power efficiency (8.0\% for speech clips and 21.4\% for silent clips) without compromising model performance (refer to Table \ref{tab:ondevice} for detailed results).

While we benchmarked using AudioSet segments with centered signals of interest, it's important to highlight that such signals are typically sparser in real-world applications. This implies that the DSL's lightweight pass usage could be more than 80\%, potentially leading to even greater reductions in on-device inference latency and power consumption for real-world use cases.

\begin{figure*}
  \centering
  \includegraphics[width=.98\linewidth]{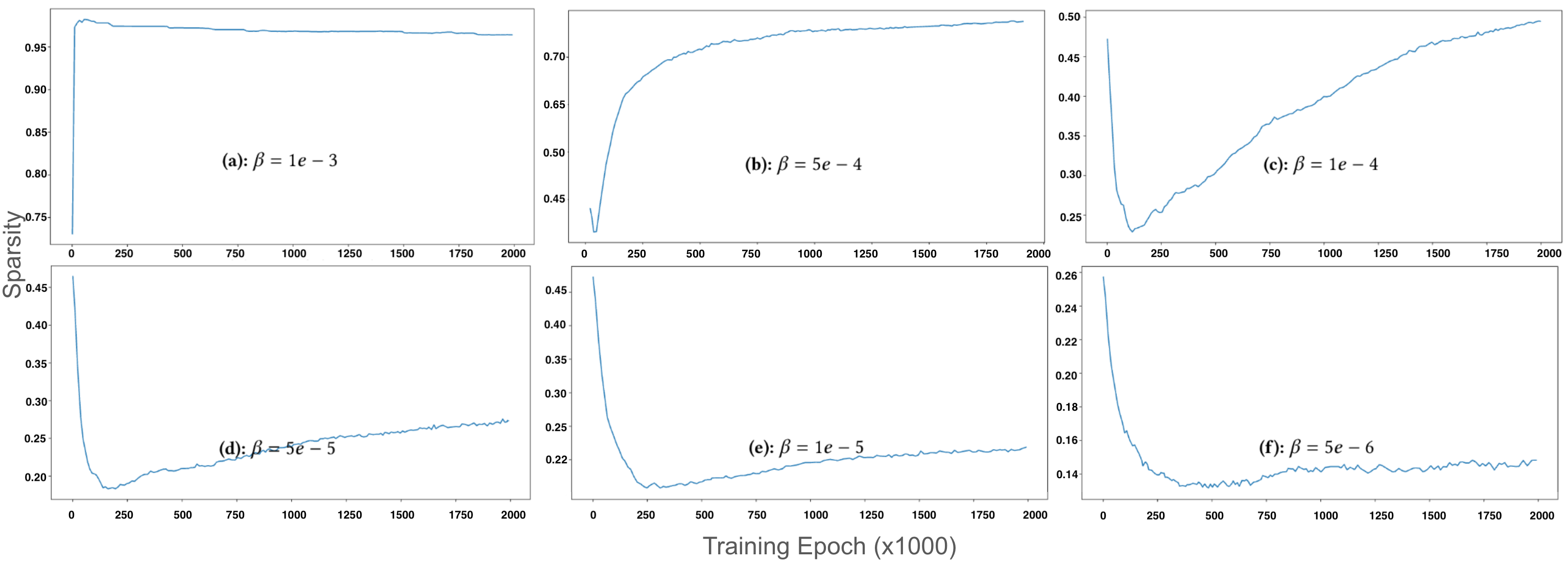}
\caption{Activation Sparsity Across Various Compression Weight $\beta$. The plots demonstrate sparsity trends over training epochs for different compression weight settings.}
  \label{fig:sparsity-beta}
\end{figure*}

\begin{figure*}
  \centering
  \includegraphics[width=.98\linewidth]{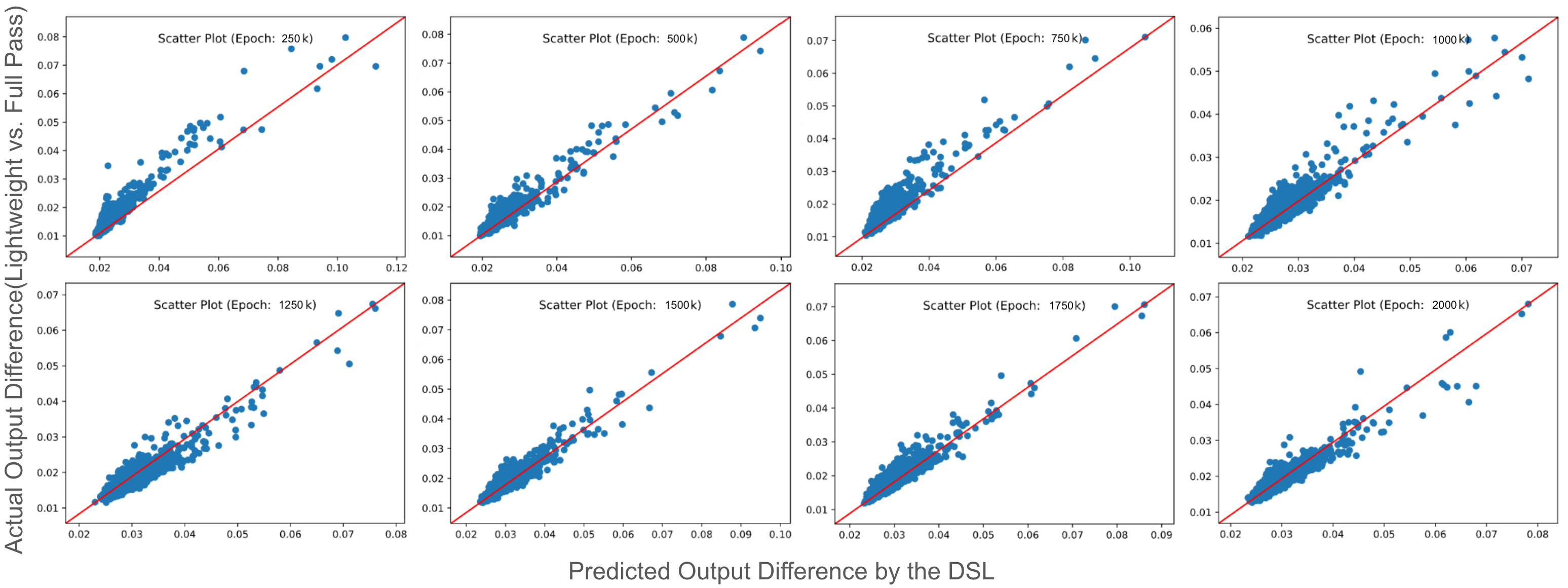}
\caption{Correlation Between Switch Predictions and Actual Differences in Encoder Embeddings. This scatter plot displays the Switch's predicted differences against the actual output differences between the original and lightweight encoder embeddings over successive training epochs.}
  \label{fig:training-switch-progress}
\end{figure*}

\begin{figure*}
  \centering
  \includegraphics[width=.98\linewidth]{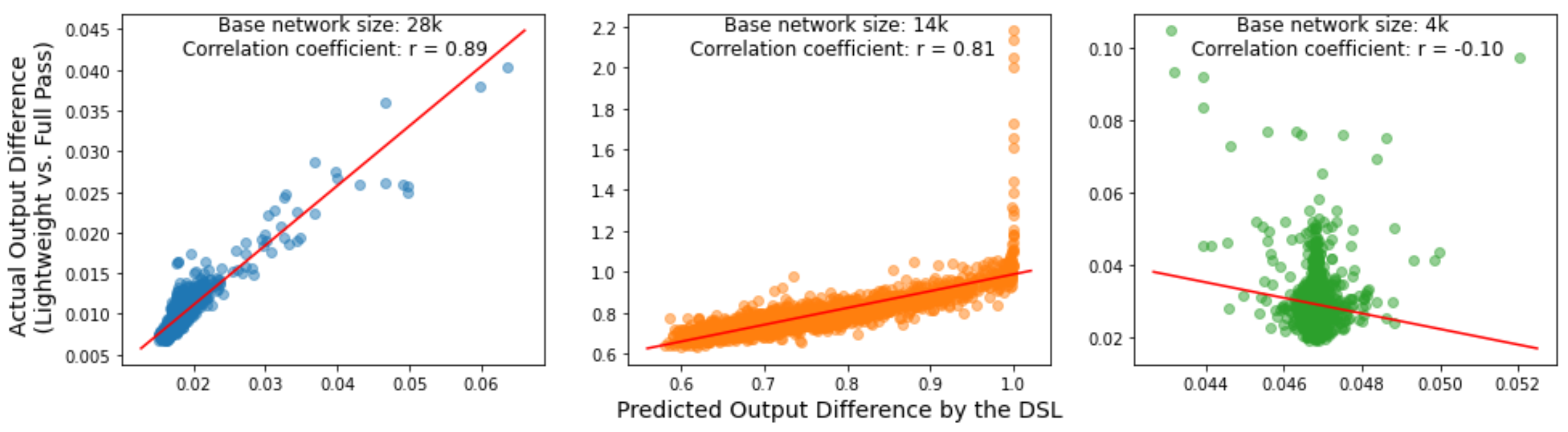}
\caption{Analysis of DSL Switch Placement and Predictive Performance. As the base network size decreases and the Switch is placed earlier, the plots exhibit a diminishing correlation between the Switch's predicted and the actual output differences.}
  \label{fig:dsl-position}
\end{figure*}

\subsubsection{Performance of Downstream Use-Cases}
To evaluate the DSL's impact on downstream use cases, a simple model (Figure \ref{fig:downstrea-usecases} (a)) was designed to directly process the SoundStream encoder output (i.e., embeddings) for tasks such as keyword detection and speaker count estimation.

As demonstrated in Figure \ref{fig:downstrea-usecases} (b) and (c), the DSL maintains downstream performance:
(1) Consistent Results: Downstream models achieve the same level of performance using DSL's mixed pass output as when using the full pass output;
(2) Potential Improvement: Interestingly, for speaker count estimation (Figure \ref{fig:downstrea-usecases} (c)), the mixed pass strategy actually outperforms the full pass. For instance, the overall accuracy for speaker count estimation using original full pass is 89.7\%, using solely lightweight pass is 94.3\%, whereas using DSL's mixed pass is 96.3\%.

The findings suggest that the DSL not only preserves the overall performance downstream but, in certain scenarios, actually enhances it. This phenomenon could be attributed to the DSL's ability to reduce over-fitting or noise in the data by simplifying the audio representations for ``easy'' segments, which is crucial for nuanced tasks like speaker count estimation. This underscores the DSL's capacity for broader network optimization, showcasing its effectiveness beyond merely maintaining baseline performance levels.

\subsection{Understanding the Compression Weight $\beta$}

The compression weight ($\beta$) in Equation \ref{eq:ssl_gcl_overall} is crucial in determining the sparsity within the activation map. Sparsity can improve computational and power efficiency.  A series of experiments investigated the impact of $\beta$ on sparsity  while ensuring that the overall performance level of the model was maintained.  Figure \ref{fig:sparsity-beta} illustrates the key findings:
(1) Sparsity Increases Over Time: As training progresses, sparsity within the activation map increases, stabilizing around epoch 1500. This suggests the model optimizes toward the desired sparsity level.
(2) $\beta$ Controls Sparsity:  As expected, larger $\beta$ values lead to greater sparsity. This allows for precise control over computational efficiency.
(3) Lower Limit on Sparsity: Interestingly, below a certain $\beta$ value (approximately 1e-5), sparsity plateaus around 15\%. This hints at an natural sparsity limit within the model's architecture.

These insights help us understand the relationship between $\beta$ and sparsity. By carefully adjusting $\beta$, we can achieve the desired balance between power efficiency and model performance, customizing the model to meet the requirements of specific applications.

\subsection{Evaluating the Switch's Performance}
Figure \ref{fig:dsl_encoder} highlights that the Switch is integral to the DSL, adeptly determining the routing for audio segments to optimize for both computational efficiency and accuracy. To analyze the Switch performance, Figure \ref{fig:training-switch-progress} plots the Switch's predicted discrepancies against the actual differences between the embeddings of the original and the lightweight encoders throughout the training epochs. 

The key findings from this analysis include: 
(1) Improved Switch Prediction Accuracy: The scatter plot's increasing alignment with the diagonal line demonstrates the Switch's growing ability to correctly identify when the lightweight pass can be used.
(2) Enhanced Lightweight Pass Performance: The decreasing MAE (Mean Absolute Error) highlights the lightweight path's increasing precision in emulating the original encoder's functionality.
(3) Challenging Cases Remain: Despite the overall performance, a subset of data points with higher MAE indicate that the full pass is still needed to achieve the best performance for certain inputs.

These observations confirm the Switch's ability to adapt over time, enhancing the system efficiency while preserving the integrity of the output. The results highlight the potential of the DSL in optimizing performance across a range of scenarios and show that the lightweight model alone cannot handle complex cases.

\newpage
\subsection{Optimizing the Switch's Placement}
Another aspect highlighted by Figure \ref{fig:dsl_encoder} is the critical role of the base network's size in shaping the accuracy of predictions made by the Switch. To delve into how the positioning of the Switch within the network impacts its ability to predict accurately, we carried out a series of experiments by varying the Switch's depth within the network architecture. The outcomes, depicted in Figure \ref{fig:dsl-position}, offer insight into the relationship between the Switch's location and its prediction efficiency. Consistent with our expectations, we observed a decline in prediction performance corresponding to reductions in the base network's size. This is because positioning the Switch too early in a base network results in insufficiently processed features, impairing its predictive accuracy. Strategic placement of the Switch is crucial for maximizing its effectiveness, as adequate feature computation is necessary for accurate Switch performance.

\section{Conclusion}
In this paper, we have presented the Dynamic Switch Layer (DSL), an innovative approach that balances computational efficiency and predictive performance. Our experiments demonstrate the DSL's ability to substantially reduce power consumption and inference latency without compromising accuracy. These advancements are achieved because of the DSL's ability to dynamically optimize neural network computation through the use of unsupervised learning.

As we advance towards sustainable AI, particularly in power-sensitive on-device applications, our work becomes increasingly relevant. Building upon Gated Compression Layers \cite{li2023gcl}, our proposed DSL facilitate in the deployment of complex machine learning tasks on resource-limited devices, expanding their potential applications.

The DSL's success highlights the value of ongoing research in on-device machine learning. We believe that this work paves the way for environmentally conscious and economically viable machine learning models that can seamlessly integrate into our daily lives.

\bibliography{paper}

\begin{thebibliography}{26}
\providecommand{\natexlab}[1]{#1}
\providecommand{\url}[1]{\texttt{#1}}
\expandafter\ifx\csname urlstyle\endcsname\relax
  \providecommand{\doi}[1]{doi: #1}\else
  \providecommand{\doi}{doi: \begingroup \urlstyle{rm}\Url}\fi

\bibitem[Abadi et~al.(2015)Abadi, Agarwal, Barham, Brevdo, Chen, Citro,
  Corrado, Davis, Dean, Devin, Ghemawat, Goodfellow, Harp, Irving, Isard, Jia,
  Jozefowicz, Kaiser, Kudlur, Levenberg, Man\'{e}, Monga, Moore, Murray, Olah,
  Schuster, Shlens, Steiner, Sutskever, Talwar, Tucker, Vanhoucke, Vasudevan,
  Vi\'{e}gas, Vinyals, Warden, Wattenberg, Wicke, Yu, and
  Zheng]{tensorflow2015-whitepaper}
Abadi, M., Agarwal, A., Barham, P., Brevdo, E., Chen, Z., Citro, C., Corrado,
  G.~S., Davis, A., Dean, J., Devin, M., Ghemawat, S., Goodfellow, I., Harp,
  A., Irving, G., Isard, M., Jia, Y., Jozefowicz, R., Kaiser, L., Kudlur, M.,
  Levenberg, J., Man\'{e}, D., Monga, R., Moore, S., Murray, D., Olah, C.,
  Schuster, M., Shlens, J., Steiner, B., Sutskever, I., Talwar, K., Tucker, P.,
  Vanhoucke, V., Vasudevan, V., Vi\'{e}gas, F., Vinyals, O., Warden, P.,
  Wattenberg, M., Wicke, M., Yu, Y., and Zheng, X.
\newblock {TensorFlow}: Large-scale machine learning on heterogeneous systems,
  2015.
\newblock URL \url{http://tensorflow.org/}.
\newblock Software available from tensorflow.org.

\bibitem[Attwell \& Laughlin(2001)Attwell and Laughlin]{attwell2001energy}
Attwell, D. and Laughlin, S.~B.
\newblock An energy budget for signaling in the grey matter of the brain.
\newblock \emph{Journal of Cerebral Blood Flow \& Metabolism}, 21\penalty0
  (10):\penalty0 1133--1145, 2001.

\bibitem[Causse et~al.(2017)Causse, Chua, Peysakhovich, Del~Campo, and
  Matton]{causse2017mental}
Causse, M., Chua, Z., Peysakhovich, V., Del~Campo, N., and Matton, N.
\newblock Mental workload and neural efficiency quantified in the prefrontal
  cortex using fnirs.
\newblock \emph{Scientific reports}, 7\penalty0 (1):\penalty0 5222, 2017.

\bibitem[Chen \& Ran(2019)Chen and Ran]{EdgeComputing:AReview2018}
Chen, J. and Ran, X.
\newblock Deep learning with edge computing: A review.
\newblock \emph{Proceedings of the IEEE}, 107\penalty0 (8):\penalty0
  1655--1674, 2019.
\newblock \doi{10.1109/JPROC.2019.2921977}.

\bibitem[Chen et~al.(2014)Chen, Du, Sun, Wang, Wu, Chen, and
  Temam]{chen2014diannao}
Chen, T., Du, Z., Sun, N., Wang, J., Wu, C., Chen, Y., and Temam, O.
\newblock Diannao: A small-footprint high-throughput accelerator for ubiquitous
  machine-learning.
\newblock \emph{ACM SIGARCH Computer Architecture News}, 42\penalty0
  (1):\penalty0 269--284, 2014.

\bibitem[Chen et~al.(2016)Chen, Krishna, Emer, and Sze]{chen2016eyeriss}
Chen, Y.-H., Krishna, T., Emer, J.~S., and Sze, V.
\newblock Eyeriss: An energy-efficient reconfigurable accelerator for deep
  convolutional neural networks.
\newblock \emph{IEEE journal of solid-state circuits}, 52\penalty0
  (1):\penalty0 127--138, 2016.

\bibitem[Chinen et~al.(2020)Chinen, Lim, Skoglund, Gureev, O'Gorman, and
  Hines]{chinen2020visqol}
Chinen, M., Lim, F. S.~C., Skoglund, J., Gureev, N., O'Gorman, F., and Hines,
  A.
\newblock Visqol v3: An open source production ready objective speech and audio
  metric, 2020.

\bibitem[Gemmeke et~al.(2017)Gemmeke, Ellis, Freedman, Jansen, Lawrence, Moore,
  Plakal, and Ritter]{jort_audioset_2017}
Gemmeke, J.~F., Ellis, D. P.~W., Freedman, D., Jansen, A., Lawrence, W., Moore,
  R.~C., Plakal, M., and Ritter, M.
\newblock Audio set: An ontology and human-labeled dataset for audio events.
\newblock In \emph{Proc. IEEE ICASSP 2017}, New Orleans, LA, 2017.

\bibitem[Grill et~al.(2020)Grill, Strub, Altch{\'e}, Tallec, Richemond,
  Buchatskaya, Doersch, Avila~Pires, Guo, Gheshlaghi~Azar,
  et~al.]{grill2020BYOL}
Grill, J.-B., Strub, F., Altch{\'e}, F., Tallec, C., Richemond, P.,
  Buchatskaya, E., Doersch, C., Avila~Pires, B., Guo, Z., Gheshlaghi~Azar, M.,
  et~al.
\newblock Bootstrap your own latent-a new approach to self-supervised learning.
\newblock \emph{Advances in neural information processing systems},
  33:\penalty0 21271--21284, 2020.

\bibitem[Gupta et~al.(2015)Gupta, Agrawal, Gopalakrishnan, and
  Narayanan]{gupta2015deep}
Gupta, S., Agrawal, A., Gopalakrishnan, K., and Narayanan, P.
\newblock Deep learning with limited numerical precision.
\newblock In \emph{International conference on machine learning}, pp.\
  1737--1746. PMLR, 2015.

\bibitem[Han et~al.(2015)Han, Mao, and Dally]{han2015deep}
Han, S., Mao, H., and Dally, W.~J.
\newblock Deep compression: Compressing deep neural networks with pruning,
  trained quantization and huffman coding.
\newblock \emph{arXiv preprint arXiv:1510.00149}, 2015.

\bibitem[He et~al.(2022)He, Chen, Xie, Li, Doll{\'a}r, and Girshick]{he2022mae}
He, K., Chen, X., Xie, S., Li, Y., Doll{\'a}r, P., and Girshick, R.
\newblock Masked autoencoders are scalable vision learners.
\newblock In \emph{Proceedings of the IEEE/CVF Conference on Computer Vision
  and Pattern Recognition}, pp.\  16000--16009, 2022.

\bibitem[Hines et~al.(2015)Hines, Skoglund, Kokaram, and Harte]{ViSQOL}
Hines, A., Skoglund, J., Kokaram, A., and Harte, N.
\newblock Visqol: an objective speech quality model.
\newblock \emph{EURASIP Journal on Audio, Speech, and Music Processing}, 2015
  (13):\penalty0 1--18, 2015.

\bibitem[Hinton et~al.(2015{\natexlab{a}})Hinton, Vinyals, and
  Dean]{Distilling2015}
Hinton, G., Vinyals, O., and Dean, J.
\newblock Distilling the knowledge in a neural network.
\newblock In \emph{NIPS Deep Learning and Representation Learning Workshop},
  2015{\natexlab{a}}.
\newblock URL \url{http://arxiv.org/abs/1503.02531}.

\bibitem[Hinton et~al.(2015{\natexlab{b}})Hinton, Vinyals, and
  Dean]{hinton2015distilling}
Hinton, G., Vinyals, O., and Dean, J.
\newblock Distilling the knowledge in a neural network.
\newblock \emph{arXiv preprint arXiv:1503.02531}, 2015{\natexlab{b}}.

\bibitem[Kingma \& Ba(2015)Kingma and Ba]{adom2015}
Kingma, D.~P. and Ba, J.
\newblock Adam: {A} method for stochastic optimization.
\newblock In Bengio, Y. and LeCun, Y. (eds.), \emph{3rd International
  Conference on Learning Representations, {ICLR} 2015, San Diego, CA, USA, May
  7-9, 2015, Conference Track Proceedings}, 2015.
\newblock URL \url{http://arxiv.org/abs/1412.6980}.

\bibitem[Li et~al.(2023)Li, Thormundsson, Poupyrev, and Gillian]{li2023gcl}
Li, H., Thormundsson, T., Poupyrev, I., and Gillian, N.
\newblock Gated compression layers for efficient always-on models, 2023.

\bibitem[Panayotov et~al.(2015)Panayotov, Chen, Povey, and
  Khudanpur]{Librispeech2015}
Panayotov, V., Chen, G., Povey, D., and Khudanpur, S.
\newblock Librispeech: An asr corpus based on public domain audio books.
\newblock In \emph{2015 IEEE International Conference on Acoustics, Speech and
  Signal Processing (ICASSP)}, pp.\  5206--5210, 2015.
\newblock \doi{10.1109/ICASSP.2015.7178964}.

\bibitem[Pervaiz et~al.(2020)Pervaiz, Vidaurre, Woolrich, and
  Smith]{pervaiz2020optimising}
Pervaiz, U., Vidaurre, D., Woolrich, M.~W., and Smith, S.~M.
\newblock Optimising network modelling methods for fmri.
\newblock \emph{Neuroimage}, 211:\penalty0 116604, 2020.

\bibitem[Sloan et~al.(2017)Sloan, Harte, Kelly, Kokaram, and
  Hines]{ViSQOLAudio}
Sloan, C., Harte, N., Kelly, D., Kokaram, A.~C., and Hines, A.
\newblock Objective assessment of perceptual audio quality using visqolaudio.
\newblock \emph{IEEE Transactions on Broadcasting}, 63\penalty0 (4):\penalty0
  693--705, 2017.
\newblock \doi{10.1109/TBC.2017.2704421}.

\bibitem[St{\"o}ter et~al.(2018)St{\"o}ter, Chakrabarty, Habets, and
  Edler]{stoter2018libricount}
St{\"o}ter, F.-R., Chakrabarty, S., Habets, E., and Edler, B.
\newblock Libricount, a dataset for speaker count estimation, 2018.

\bibitem[Tan et~al.(2019)Tan, Chen, Pang, Vasudevan, Sandler, Howard, and
  Le]{tan2019mnasnet}
Tan, M., Chen, B., Pang, R., Vasudevan, V., Sandler, M., Howard, A., and Le,
  Q.~V.
\newblock Mnasnet: Platform-aware neural architecture search for mobile.
\newblock In \emph{Proceedings of the IEEE/CVF Conference on Computer Vision
  and Pattern Recognition}, pp.\  2820--2828, 2019.

\bibitem[Warden(2018)]{warden2018speech}
Warden, P.
\newblock Speech commands: A dataset for limited-vocabulary speech recognition,
  2018.

\bibitem[Zeghidour et~al.(2021)Zeghidour, Luebs, Omran, Skoglund, and
  Tagliasacchi]{zeghidour2021soundstream}
Zeghidour, N., Luebs, A., Omran, A., Skoglund, J., and Tagliasacchi, M.
\newblock Soundstream: An end-to-end neural audio codec.
\newblock \emph{IEEE/ACM Transactions on Audio, Speech, and Language
  Processing}, 30:\penalty0 495--507, 2021.

\bibitem[Zen et~al.(2019)Zen, Dang, Clark, Zhang, Weiss, Jia, Chen, and
  Wu]{zen2019libritts}
Zen, H., Dang, V., Clark, R., Zhang, Y., Weiss, R.~J., Jia, Y., Chen, Z., and
  Wu, Y.
\newblock Libritts: A corpus derived from librispeech for text-to-speech, 2019.

\bibitem[Zhou et~al.(2017)Zhou, Yao, Guo, Xu, and Chen]{zhou2017incremental}
Zhou, A., Yao, A., Guo, Y., Xu, L., and Chen, Y.
\newblock Incremental network quantization: Towards lossless cnns with
  low-precision weights.
\newblock \emph{arXiv preprint arXiv:1702.03044}, 2017.

\end{thebibliography}
\bibliographystyle{icml2023}
\end{document}